\documentclass[a4paper,10pt, conference]{IEEEtran}

\IEEEoverridecommandlockouts

\usepackage[a4paper,left=62pt,right=62pt,top=78pt,bottom=78pt]{geometry}

\usepackage{cite}
\usepackage{amsmath, amssymb, amsfonts}
\usepackage{graphicx}
\usepackage{tikz}
\usepackage{textcomp}
\usepackage{multirow}

\usepackage{algpseudocode}
\usepackage{algorithm}

\usepackage{subcaption}
\usepackage{url}
\usepackage{lipsum}
\usepackage[utf8]{inputenc}

\begin{document}

\title{ORPA: Online Residual Policy Adaptation for Robot Manipulation Control with Human Feedback\\
}

\author{
\begin{tabular}{cccc}
\textbf{Muhammad A. Muttaqien$^{*}$} &
\textbf{Tomohiro Motoda$^{*}$} &
\textbf{Ryo Hanai} &
\textbf{Yukiyasu Domae}
\end{tabular}
\\[4pt]
\textit{Embodied AI Research Team, National Institute of AIST, Tokyo, Japan}
\\[2pt]
\small
muha.muttaqien@aist.go.jp \quad
tomohiro.motoda@aist.go.jp \quad
ryo.hanai@aist.go.jp \quad
domae.yukiyasu@aist.go.jp
}

\maketitle
\footnotetext[1]{These authors contributed equally to this work.}

\begin{abstract}
Robotic manipulation policies trained via imitation learning, such as Action Chunking with Transformers (ACT), can achieve strong performance under ideal conditions but often remain sensitive to small execution errors and distribution shifts. Correcting these failures typically requires dataset aggregation and full-policy retraining, which is computationally expensive and unsuitable for real-time deployment. In this work, we propose Online Residual Policy Adaptation (ORPA), a framework that enables immediate, feedback-driven correction of robot actions without modifying the underlying policy parameters. ORPA augments a pretrained control policy with a lightweight, feedback-conditioned module that predicts residual adjustments directly in joint space, allowing the system to adapt its behavior at runtime. We evaluate ORPA on a set of precision-sensitive manipulation tasks using the ALOHA platform, demonstrating improvements in success rate and recovery from small perturbations compared to baseline control policies and rule-based inverse kinematics corrections.
\end{abstract}


\section{Introduction}
Recent advances in imitation learning have enabled robotic manipulation systems to achieve impressive performance across a wide range of tasks \cite{Torabi2018, Chi2023}. In particular, transformer-based policies such as Action Chunking with Transformers (ACT) have demonstrated strong capabilities in learning long-horizon manipulation behaviors directly from human demonstrations. Combined with low-cost bimanual robotic platforms such as ALOHA, these approaches have accelerated research toward scalable and accessible robot learning systems. Despite these advances, imitation learning policies often remain highly sensitive to small execution errors, environmental variations, and distribution shifts encountered during real-world deployment.

In practical robotic manipulation scenarios, even small deviations in end-effector position, orientation, or timing can lead to task failure. This issue becomes particularly critical in precision-sensitive tasks such as cluttered object grasping, narrow-space pick-and-place, and coordinated bimanual manipulation, where minor spatial offsets may cause collisions, unstable grasps, or failed object transfers. Although pretrained policies may achieve high success rates under nominal conditions, their performance can degrade when objects are slightly displaced, viewpoints change, or execution conditions differ from the training distribution.

A common approach to improving policy robustness is to collect additional demonstrations and retrain or fine-tune the policy using dataset aggregation techniques \cite{Ross2011, Lee2017, Bi2020}. While effective, retraining-based pipelines are computationally expensive, require repeated data collection, and are often unsuitable for interactive or real-time robotic deployment. Moreover, these approaches typically modify the entire policy network even when failures originate only from small local execution errors. Another straightforward alternative is to apply rule-based geometric corrections using forward and inverse kinematics, where feedback signals such as “too left” or “too high” are translated into predefined end-effector offsets and projected back into joint space. Although intuitive, such methods assume that identical feedback always corresponds to identical corrective behavior, despite the fact that appropriate corrections often depend on task phase, object configuration, and execution context. Consequently, both retraining-based adaptation and fixed rule-based correction methods struggle to provide efficient and flexible real-time adaptation.

Instead of modifying the underlying policy parameters, ORPA augments a pretrained control policy with a feedback-conditioned residual module that predicts corrective action adjustments directly in joint space. Given the current policy output and external feedback, the proposed module produces temporally consistent residual corrections that refine robot behavior during execution while preserving the original policy performance. Unlike retraining-based approaches, ORPA enables immediate adaptation without requiring additional optimization of the base policy. Furthermore, by learning residual corrections rather than relying on predefined geometric rules, the proposed framework can capture context-dependent and behavior-level adjustments that extend beyond simple end-effector offsets.

We evaluate ORPA on a set of precision-sensitive robotic manipulation tasks using the ALOHA platform. Our experiments focus on scenarios where small execution errors significantly affect task success, including spatial perturbations and coordination-sensitive bimanual tasks. Experimental results demonstrate that ORPA improves task robustness and recovery performance compared to baseline ACT policies and rule-based inverse kinematics correction methods. 

\begin{figure}[t]
    \centering
    \includegraphics[width=\columnwidth]{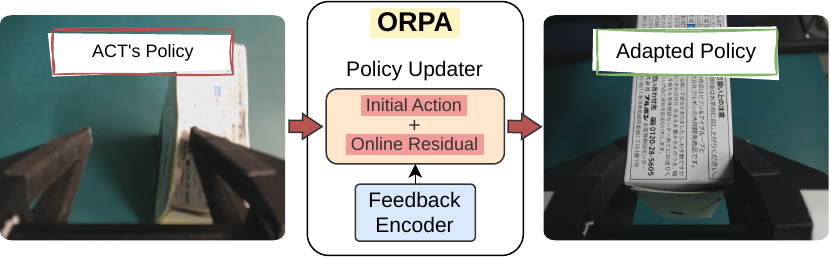}
    \caption{Overview of proposed ORPA. Online residual corrections generated from human feedback refine pretrained ACT actions for robust manipulation.}
    \label{fig:simplediagram}
% \vspace{-4mm}
\end{figure}

\section{RELATED WORK}
Imitation learning has become one of the dominant paradigms for robotic manipulation due to its ability to learn complex behaviors directly from human demonstrations \cite{Zare2024}. Recent advances in transformer-based architectures have significantly improved the capability of manipulation policies to model sequential actions and multimodal observations. In particular, Action Chunking with Transformers (ACT) introduced action chunking strategies for efficient and stable policy learning in bimanual robotic manipulation tasks. Combined with platforms such as ALOHA, these approaches have enabled scalable collection of manipulation demonstrations and strong performance across diverse tabletop tasks. However, despite their effectiveness, imitation learning policies often remain sensitive to small distribution shifts and execution perturbations encountered during deployment.

Several studies have explored methods for improving policy robustness through online adaptation and iterative refinement. A common strategy is dataset aggregation and policy retraining, where new demonstrations or corrective samples are incorporated to improve future performance. Representative approaches such as DAgger \cite{Ross2011} and HG-Dagger \cite{Kelly2019} iteratively collect corrective supervision to reduce compounding errors during policy execution. More recent methods have investigated online policy adaptation for robotic manipulation through feedback-driven updates and continual learning mechanisms. However, many of these approaches require expensive optimization procedures, repeated retraining, or modification of the underlying policy parameters, limiting their applicability in real-time robotic systems.

Among recent works, OLAF \cite{Liu2023} and YAY Robot \cite{Shi2024} introduced an interactive learning framework that incorporates human verbal corrections for robotic manipulation. OLAF demonstrates the effectiveness of language-guided corrective supervision and offline policy refinement for improving robotic behavior through iterative data collection and fine-tuning. Another common strategy for handling manipulation errors is the use of rule-based geometric correction through forward and inverse kinematics, where human feedback or task errors are mapped into predefined end-effector offsets and converted back into joint-space actions using inverse kinematics solvers. While such approaches are intuitive and computationally efficient for simple local adjustments, they assume that identical feedback always corresponds to identical corrective behavior. In practice, however, appropriate corrections often depend on the task phase, object configuration, robot pose, and execution context. For example, the same feedback signal such as “too left” may require different correction magnitudes during object approach, grasping, or placement stages.

In contrast to prior approaches, ORPA combines the efficiency of lightweight online adaptation with the flexibility of learning-based residual correction. By conditioning corrective actions on human feedback while preserving the pretrained manipulation policy, ORPA enables real-time behavioral refinement without requiring full-policy retraining or manually designed correction rules.

\begin{figure*}[tbp]
% \vspace{-20mm}
  \centering
  \includegraphics[width=\linewidth]{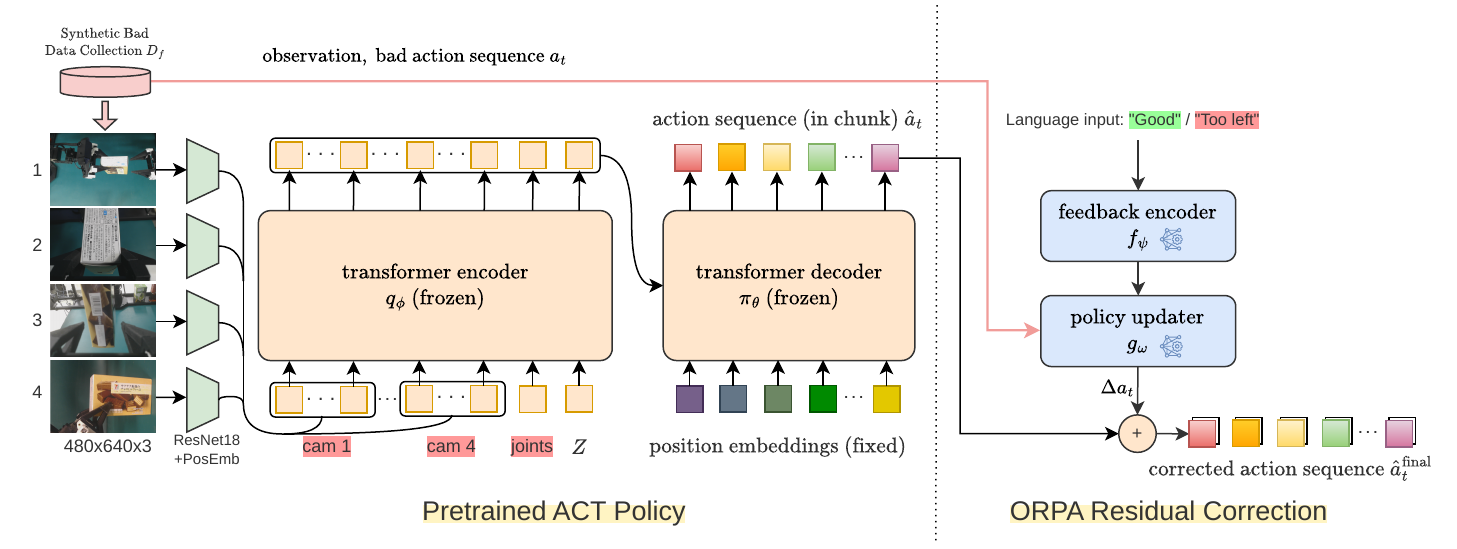}
  \caption{Training pipeline of the proposed Online Residual Policy Adaptation (ORPA) framework. \textit{Left}: A pretrained ACT policy generates nominal joint-space actions from visual observations and robot states. Structured perturbations are introduced to create synthetic errors together with corresponding corrective feedback labels. \textit{Right}: The Feedback Encoder (FE) and Policy Updater (PU) are trained to predict residual action corrections conditioned on RGB images, robot states, bad actions, and human feedback. The predicted residual is supervised using the difference between perturbed and reference actions.}
  \label{fig:orpa_diagram}
% \vspace{-2mm}
\end{figure*}

\section{EXPERIMENT SETUP}
The proposed Online Residual Policy Adaptation (ORPA) framework, shown in Figure~\ref{fig:simplediagram}, was evaluated in both simulation and real-world environments using the ALOHA platform. The experimental setup consists of a dual-arm bimanual manipulation system with vision and joint-state information, along with precision manipulation tasks that require coordinated control and tight tolerance specifications. This section describes the experimental setup used throughout the experiments.

\subsection{ALOHA Simulation}

Simulation experiments were conducted using a MuJoCo-based \cite{Todorov2012} ALOHA simulation environment integrated with the ACT framework. The simulator models a synchronized dual-arm manipulation system consisting of two 6-DOF manipulators with parallel-jaw grippers, resulting in a 14-dimensional joint-space action representation including gripper actuation. The environment operates at a control frequency of 50 Hz.

\subsection{ALOHA Workspace}

The real-world experiments were conducted using the ALOHA platform, a dual-arm robotic manipulation system designed for bimanual imitation learning and manipulation research. The system configuration consists of two collaborative manipulators, parallel-jaw grippers, and a multi-view RGB vision setup, as described below.

\begin{itemize}
    \item \textbf{Robotic Arm:} The real-world robotic platform is based on the ALOHA system consisting of two synchronized 6-DOF manipulators configured for bimanual manipulation tasks. Both arms are controlled through a ROS-based \cite{Quigley2009} joint-space interface operating at 50 Hz, providing a combined 14-dimensional control space including gripper actuation. ACT predicts chunked joint trajectories while ORPA generates online residual joint-space corrections during execution.

    \item \textbf{Gripper:} Each manipulator is equipped with a parallel-jaw gripper supporting variable-width grasping for both pinch and encompassing grasps. The grippers support real-time actuation control and are integrated directly into the policy action representation for coordinated bimanual manipulation.

    \item \textbf{Vision System:} The platform utilizes four synchronized RGB cameras, including two wrist-mounted cameras and two external tabletop cameras, providing multi-view observations of the workspace. All RGB images are resized to $640 \times 480$ resolution and directly processed by the ACT visual encoder without explicit object detection modules.
\end{itemize}

\section{ONLINE RESIDUAL POLICY ADAPTATION}

The proposed Online Residual Policy Adaptation (ORPA) framework builds upon a pretrained Action Chunking with Transformers (ACT) policy as the underlying manipulation controller. ACT is an imitation learning framework that predicts temporally coherent chunks of joint-space actions from multimodal observations, including multi-view RGB images and robot joint states. By predicting action sequences rather than individual low-level commands, ACT enables smooth and stable execution of long-horizon manipulation tasks.

In this work, ACT serves as a frozen base policy and is not modified during ORPA training or deployment. Given the current observation $o_t$, ACT predicts a chunk of future actions $\hat{a}_{t:t+k}$, which provides the nominal manipulation behavior. While ACT demonstrates strong performance under nominal conditions, its performance can degrade when execution errors, object perturbations, or distribution shifts occur during deployment. ORPA addresses this limitation by introducing a feedback-conditioned residual adaptation module that refines ACT actions online without requiring retraining of the underlying policy, making it well suited for real-time correction.

\subsection{ORPA Architecture}
Figure~\ref{fig:orpa_diagram} illustrates the overall architecture of the proposed ORPA framework. The framework consists of a pretrained ACT policy, a Feedback Encoder, and a Policy Updater module. During execution, ACT first generates a nominal action chunk $\hat{a}_{t:t+k}$ based on the current visual observations and robot states. Simultaneously, external corrective feedback provided by a human operator is encoded into a compact latent representation. The encoded feedback is combined with the current robot state, visual observations, and ACT predictions to estimate a residual action correction $\Delta a_t$. Instead of replacing the original ACT output, ORPA performs residual adaptation by adding the predicted correction to the ACT action:

\begin{equation}
a^{final}_{t:t+k} = \hat{a}_{t:t+k} + \Delta a_t
\end{equation}

where $\hat{a}_{t:t+k}$ denotes the nominal ACT prediction and $\Delta a_t$ represents the feedback-conditioned residual correction. This residual formulation preserves the original manipulation behavior learned by ACT while allowing local corrective adjustments during execution. Since ORPA operates directly in joint space, corrective actions can be applied without requiring explicit forward or inverse kinematics computations.

\subsection{Feedback Encoder and Policy Updater}

To enable online corrective adaptation, ORPA introduces two lightweight transformer-based components, a Feedback Encoder and a Policy Updater. The Feedback Encoder converts human corrective feedback into a latent representation suitable for policy adaptation. During training and evaluation, feedback is represented as either discrete or continuous signals. Discrete feedback uses categorical commands such as \textit{too left}, \textit{too right}, \textit{too high}, and \textit{good}, while continuous feedback numerically encodes both correction direction and magnitude. These feedback signals are provided by human observers immediately after the robot's initial attempt fails to successfully execute the task. Each feedback signal is first represented as a discrete token and then mapped into a learnable embedding space. Let $f_t$ denote the feedback signal at timestep $t$. The Feedback Encoder produces a latent feedback representation:

\begin{equation}
r_t = f_\psi(f_t)
\end{equation}

where $r_t$ captures the semantic meaning of the corrective instruction. The Policy Updater receives the encoded feedback representation together with the current robot observation and ACT prediction. The module predicts a residual action correction:

\begin{equation}
\Delta a_t = g_\omega(r_t, o_t, {a}_{t:t+k})
\end{equation}

where $o_t$ denotes the current observation and $a_{t:t+k}$ denotes the action trajectory associated with the current manipulation context. By conditioning residual actions on both feedback and execution context, ORPA can generate context-dependent corrections whose magnitude and direction vary according to the task stage, object configuration, and robot state. This differs from rule-based inverse kinematics approaches, which apply identical corrections regardless of execution context.

\subsection{Synthetic Data Generation}

\begin{figure}[tbp]
    \centering
    \begin{tikzpicture}
        \node[inner sep=0pt, rounded corners=20pt] {\includegraphics[width=\linewidth]{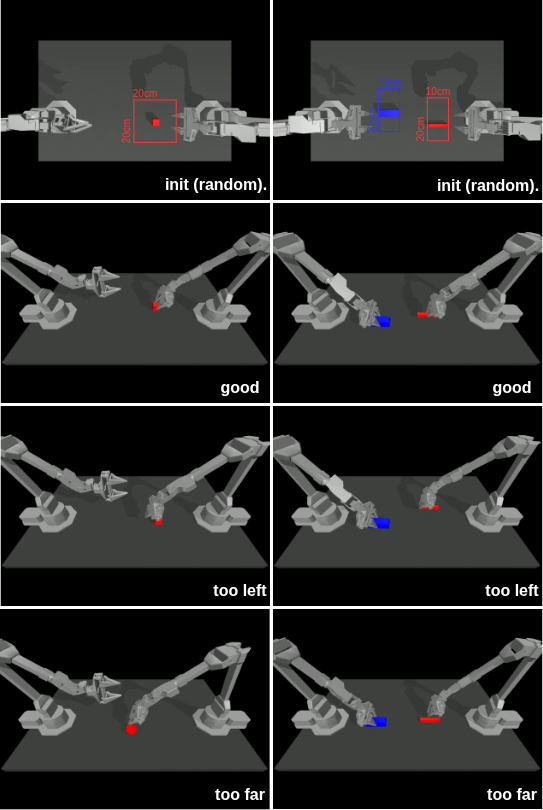}};
    \end{tikzpicture}
    \caption{Generation of synthetic manipulation errors for ORPA training. Objects are randomly initialized within predefined workspace boundaries (top row), and successful demonstrations are first collected as reference trajectories (good). Cartesian perturbations are then introduced to the target object pose, producing failure scenarios (too left and too far).}
    \label{fig:data_generation}
    % \vspace{-4mm}
\end{figure}

Collecting large-scale corrective demonstrations from human operators can be expensive and time-consuming. To address this limitation, ORPA generates corrective training samples automatically from successful demonstration trajectories. As illustrated in Figure~\ref{fig:data_generation}, successful manipulation trajectories are first collected using teleoperation and used to train the ACT policy. Controlled perturbations are then introduced into the manipulation environment by modifying object positions and orientations within predefined ranges. These perturbations create failure scenarios that mimic common execution errors encountered during deployment.

For each perturbation, a corresponding feedback label is assigned according to the perturbation direction. For example, leftward object displacements are labeled as \textit{too left}, while upward displacements are labeled as \textit{too high}. The resulting dataset consists of perturbed observations, corrective feedback labels, and corresponding reference actions obtained from the original successful demonstrations. The residual correction target is computed as the difference between the reference action and the perturbed action:

\begin{equation}
\Delta a_t^{*} = a_t^{ref} - a_t^{perturbed}
\end{equation}

where $a_t^{ref}$ denotes the action from the successful trajectory and $a_t^{perturbed}$ denotes the action associated with the perturbed execution. This procedure enables scalable generation of corrective supervision without requiring additional teleoperation demonstrations.

\subsection{Training Objective}

ORPA is trained while keeping the ACT policy fixed. During training, the pretrained ACT policy first predicts a nominal action chunk based on the current observation. The Feedback Encoder and Policy Updater then estimate a residual correction conditioned on the feedback signal and execution context. The final corrected action is computed as:

\begin{equation}
a^{final}_{t:t+k}
=
\hat{a}_{t:t+k}
+
\Delta a_t
\end{equation}

The residual adaptation module is optimized using a mean-squared error objective between the corrected action and the reference action:

\begin{equation}
L_{ORPA}
=
\frac{1}{N}
\sum_{i=1}^{N}
\left\|
a^{final}_i
-
a^{ref}_i
\right\|^2
\end{equation}

where $a^{ref}_i$ denotes the target action obtained from the original successful demonstration. The complete training procedure is summarized in Algorithm~\ref{alg:orpa_train}. Since only the Feedback Encoder and Policy Updater are optimized, ORPA introduces minimal computational overhead while preserving the original ACT policy parameters.

\begin{algorithm}[t]
\caption{ACT Training}
\label{alg:act_train}
\begin{algorithmic}[1]
\State \textbf{Given:} Demo dataset $\mathcal{D}$, chunk size $k$, weight $\beta$.
\State Let $a_t, o_t$ represent action and observation at timestep $t$, $\bar{o}_t$ represent $o_t$ without image observations.
\State Initialize encoder $q_\phi(z \mid a_{t:t+k}, \bar{o}_t)$
\State Initialize decoder $\pi_\theta(\hat{a}_{t:t+k} \mid o_t, z)$
\For{iteration $n = 1,2,\dots$}
    \State Sample $o_t, a_{t:t+k}$ from $\mathcal{D}$
    \State Sample $z$ from $q_\phi(z \mid a_{t:t+k}, \bar{o}_t)$
    \State Predict $\hat{a}_{t:t+k}$ from $\pi_\theta(\hat{a}_{t:t+k} \mid o_t, z)$
    \State $\mathcal{L}_{\text{reconst}} = MSE(\hat{a}_{t:t+k}, a_{t:t+k})$
    \State $\mathcal{L}_{\text{reg}} = D_{KL}(q_\phi(z \mid a_{t:t+k}, \bar{o}_t)\,\|\,\mathcal{N}(0,I))$
    \State Update $\theta, \phi$ with ADAM and $\mathcal{L} = \mathcal{L}_{\text{reconst}} + \beta \mathcal{L}_{\text{reg}}$
\EndFor
\end{algorithmic}
\end{algorithm}

\begin{algorithm}[t]
\label{alg:act_orpa}
\caption{ORPA Training with Pretrained ACT Policy}
\label{alg:orpa_train}
\begin{algorithmic}[1]

\State \textbf{Given:} Pretrained ACT policy $\pi_\theta$, feedback dataset $\mathcal{D}_f$.

\State Initialize feedback encoder $f_\psi$.

\State Initialize policy updater $g_\omega$.

\For{iteration $n = 1,2,\dots$}

    \State Sample $(o_t, f_t, a_{t:t+k})$ from $\mathcal{D}_f$

    \State Predict ACT action chunk:
    $\hat{a}_{t:t+k} = \pi_\theta(o_t)$

    \State Encode feedback signal:
    $r_t = f_\psi(f_t)$

    \State Predict residual correction:
    $\Delta a_t = g_\omega(r_t, o_t, {a}_{t:t+k})$

    \State Apply residual correction:
    $\hat{a}^{\text{final}}_{t:t+k} =
    \hat{a}_{t:t+k} + \Delta a_t$

    \State Compute residual loss:
    $\mathcal{L}_{\text{ORPA}} =
    MSE(\hat{a}^{\text{final}}_{t:t+k}, a_{t:t+k})$

    \State Update $\psi, \omega$ with ADAM

\EndFor

\end{algorithmic}
\end{algorithm}

\section{EXPERIMENTS AND ANALYSIS}

\subsection{Tasks}

We evaluate the proposed ORPA framework on several precision-sensitive bimanual manipulation tasks using the ALOHA platform \cite{Zhao2023, Fu2024, Aldaco2024}. The evaluation tasks include precision pick-and-place, cluttered object grasping, constrained placement, and coordinated bimanual manipulation scenarios. These tasks were selected because small spatial or temporal deviations can significantly affect task success. During evaluation, object poses were randomly perturbed relative to demonstration trajectories to simulate realistic distribution shifts and execution errors. Corrective feedback signals such as “too left,” “too right,” and “too high” were provided online during task execution to evaluate the effectiveness of residual action-level adaptation.

\begin{table}[t]
\centering
\caption{Hyperparameters used for ORPA training.}
\label{tab:hyperparams}

\renewcommand{\arraystretch}{1.2}
\setlength{\tabcolsep}{10pt}

\begin{tabular}{ll}
\hline

\textbf{ORPA Training} \\[4pt]
\hline
Batch Size & 8 \\
Chunk Size & 100 \\
KL Weight & 10 \\
Learning Rate & 1e-5 \\
Number of Epochs & 2000 \\

\hline

\textbf{ORPA (Feedback Encoder)} \\[4pt]
\hline
Embedding Dimension & 32 \\
Hidden Dimension & 128 \\
Output Dimension & 64 \\
Learning Rate & 1e-5 \\

\hline

\textbf{ORPA (Policy Updater)} \\[4pt]
\hline
Action Dimension & 14 \\
Feedback Dimension & 64 \\
Hidden Dimension & 128 \\
Number of Attention Heads & 6 \\
Number of Layers & 2 \\
Learning Rate & 1e-5 \\

\hline
\end{tabular}
% \vspace{-2mm}
\end{table}

\subsection{ACT Hyper-parameters Tuning}

The ACT policy was trained using chunked joint-space action prediction with a Transformer-based architecture. The action chunk size was set to 100 control steps with a policy update frequency of 50\,Hz. Multi-view RGB observations with resolution $640 \times 480$ were encoded using convolutional visual backbones before fusion with proprioceptive robot states. Hyperparameters were selected based on validation success rate and trajectory stability during manipulation execution. The detailed hyperparameter settings are summarized in Table \ref{tab:hyperparams}.

\subsection{Experiment Results}

Our experiments demonstrate that the proposed ORPA framework improves manipulation robustness under small execution perturbations and distribution shifts compared to baseline ACT policies and rule-based inverse kinematics (IK) correction methods. As shown in Table \ref{tab:failure_comparison}, ORPA consistently achieved higher success rates across all evaluated tasks. For the Cube Transfer task, the original ACT policy achieved 60.0\% success under failure conditions, while ORPA improved performance to 92.3\% and 91.7\% using discrete and continuous feedback, respectively. Similar improvements were observed in the Bimanual Insertion task, where success rates increased from 60.0\% to 80.3\% and 84.7\%.

The Bimanual Insertion task proved more challenging than Cube Transfer due to its tighter geometric constraints. While Cube Transfer only requires the receiving gripper to securely hold the cube, successful insertion requires accurate peg-target alignment and physical contact, making the task significantly less tolerant to small positional, rotational, and timing errors. Nevertheless, ORPA maintained substantial performance gains over the baseline ACT policy, demonstrating its effectiveness in precision-sensitive manipulation scenarios.

ORPA also consistently outperformed the rule-based IK baseline. Fixed IK assumes that identical feedback should produce identical corrective motion, which can cause over- or under-correction when the actual error magnitude varies. In contrast, ORPA predicts feedback-conditioned residual actions directly from visual inputs, robot states, and policy outputs, enabling context-dependent corrections without forward- and inverse-kinematics conversion during execution. Rule-based IK also lacks task-phase awareness, applying the same correction during approach, grasping, lifting, or transfer. By learning from demonstration trajectories, ORPA implicitly captures task context and can adjust correction strength accordingly. Results with combined translational and rotational feedback further demonstrate the flexibility of the approach, although rotational corrections remain more challenging due to additional orientation constraints.

\subsection{Comparison with OLAF}

We additionally compare ORPA with OLAF, a language-based correction framework that incorporates human feedback through large language model (LLM) reasoning. When adapted for online correction, OLAF achieves a success rate of 78.5\%, demonstrating the effectiveness of feedback-guided manipulation recovery. However, its performance remains below the proposed ORPA framework. A key limitation of OLAF for real-time deployment is inference latency, as each correction requires an additional LLM inference step. Furthermore, corrective decisions are generated at individual execution steps rather than directly over action trajectories, which may reduce temporal consistency during manipulation. OLAF also requires a forward-kinematics and inverse-kinematics conversion pipeline to translate feedback into executable actions. In contrast, ORPA predicts residual corrections directly in joint space and operates on chunked action trajectories generated by ACT. 

\begin{table}[t]
\centering
\caption{
Performance under failure conditions in simulation. All methods achieve 90.0\% (Cube Transfer) and 84.0\% (Bimanual Insertion) success under normal conditions.
}
\label{tab:failure_comparison}

\renewcommand{\arraystretch}{1.15}

\resizebox{\columnwidth}{!}{%
\begin{tabular}{l|c|c|c}
\hline
\textbf{Method} & \textbf{Feedback} & \textbf{Data Type} & \textbf{Failures} \\
\hline

\multicolumn{4}{c}{\textbf{Cube Transfer (Sim)}} \\[3pt]
\hline

Original ACT & Translation & -- & 60.0\% \\
ACT + ORPA (Ours) & Translation & Discrete & \textbf{92.3\%} \\
ACT + ORPA (Ours) & Translation & Continuous & \textbf{91.7\%} \\
ACT + ORPA (Ours) & Trans. + Rotation & Discrete & \textbf{85.9\%} \\
ACT + Rule-based IK & Translation & Discrete & 81.0\% \\
ACT + Rule-based IK & Translation & Continuous & 76.0\% \\
ACT + OLAF & Translation & -- & 78.5\% \\

\hline
\multicolumn{4}{c}{\textbf{Bimanual Insertion (Sim)}} \\[3pt]
\hline

Original ACT & Translation & -- & 60.0\% \\
ACT + ORPA (Ours) & Translation & Discrete & \textbf{80.3\%} \\
ACT + ORPA (Ours) & Translation & Continuous & \textbf{84.7\%} \\

\hline
\end{tabular}%
}
\end{table}

\section{REAL-WORLD VALIDATION}

\begin{figure*}[tbp]
% \vspace{-20mm}
  \centering
  \includegraphics[width=\linewidth]{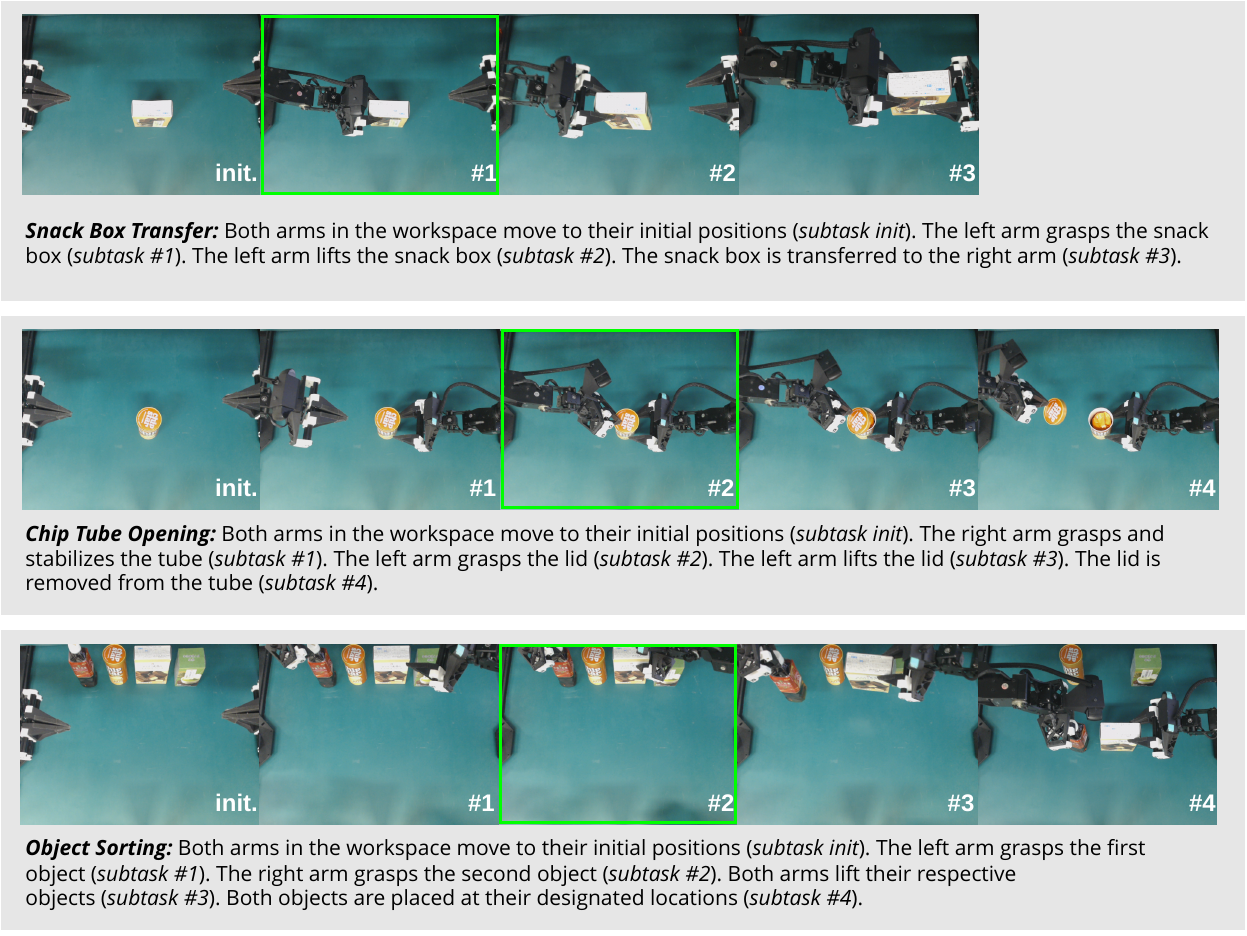}
  \caption{Qualitative results of the proposed ORPA framework on three real-world manipulation tasks: snack box transfer, chip tube opening, and object sorting. Each sequence illustrates the initialization stage followed by key manipulation subtasks. The ones with green boxes indicate the subtasks during which online adaptation usually occurs.}
  \label{fig:qualitative_results}
% \vspace{-2mm}
\end{figure*}

\subsection{Deployment Setup}

To evaluate the practical applicability of the proposed ORPA framework, we deployed the system on the real-world ALOHA robotic platform. Experiments were conducted on three precision-sensitive manipulation tasks: snack box transfer, chip tube opening, and object sorting. These tasks require accurate grasping, coordinated motion, and precise object interaction, making them suitable for evaluating online corrective adaptation under real-world conditions. The original ACT policy was trained using 100 successful teleoperated demonstrations for each task. The trained ACT policy exhibited strong manipulation capabilities and was able to perform limited recovery behaviors in certain situations. For example, when object grasping initially failed, the policy occasionally re-attempted grasping and successfully completed the task. For training the ORPA modules, namely the Feedback Encoder and Policy Updater, corrective data were generated using two strategies: (1) action perturbation only, and (2) action perturbation combined with perturbed observations.

\subsection{Task Evaluation}

Table~\ref{tab:realworld_results} summarizes the real-world performance of the original ACT policy and the proposed ORPA variants on three manipulation tasks: snack box transfer, chip tube opening, and object sorting. Task performance was evaluated over 30 trials for each task using a task completion score, where full success received a score of 1.0 and partial completion received a score of 0.5. Partial scores were assigned to intermediate outcomes, such as successful grasping without successful placement. The results show that the original ACT policy already achieves strong performance across all tasks. This observation is consistent with qualitative findings that ACT exhibits a degree of robustness and can occasionally recover from execution failures, such as re-attempting object grasping after an initial miss. Such behaviors suggest that ACT possesses both interpolation capability within the training distribution. 

\begin{table}[h!]
\centering
\caption{
Real-world task performance. Partial task completion receives a score of 0.5, while complete task success receives a score of 1.0. Note that Original ACT is evaluated under normal conditions, without induced failures.
}
\label{tab:realworld_results}

\renewcommand{\arraystretch}{1.25}

\begin{tabular}{l|c|c|c}
\hline
\textbf{Method} & \textbf{Transfer} & \textbf{Open} & \textbf{Sort} \\
\hline

Original ACT
& 0.83
& 0.79
& 0.80 \\

ACT + ORPA (Action Only)
& 0.85
& 0.81
& 0.81 \\

ACT + ORPA (Action + Observation)
& \textbf{0.86}
& \textbf{0.82}
& \textbf{0.83} \\

\hline
\end{tabular}

\end{table}

Nevertheless, the chip tube opening task remains particularly challenging due to its contact-rich interactions, precise alignment requirements, and sensitivity to timing errors, resulting in the lowest baseline performance among the evaluated tasks. Introducing ORPA with action perturbation alone improves the task scores, demonstrating that residual policy adaptation can effectively correct execution failures during deployment. Furthermore, incorporating perturbed observations further enhances robustness and improves performance across all tasks. These results indicate that a small amount of additional observation-level correction data can significantly improve robustness while maintaining data efficiency. Importantly, real-world failures are not limited to translation and rotation errors. Additional failure modes include timing mismatches, imperfect contact interactions, and execution inconsistencies arising from object dynamics. Consequently, ORPA is designed to perform corrective adaptation based on the provided feedback signal rather than assuming predefined categories of failure.

\subsection{Qualitative Results}

Figure~\ref{fig:qualitative_results} presents representative execution sequences of the proposed ORPA framework on three real-world manipulation tasks: snack box transfer, chip tube opening, and object sorting. Each sequence illustrates the robot's behavior under online corrective feedback, highlighting both failure recovery and successful task completion. Real-world experiments demonstrate that ORPA can effectively improve manipulation robustness through online corrective feedback. When the robot encountered execution failures, feedback signals such as \textit{too left}, \textit{too right}, or \textit{too high} generated residual corrections that refined the action trajectory without modifying the underlying ACT policy. Unlike fixed rule-based correction methods, ORPA adapts its corrective behavior according to the current execution context and task phase.

\begin{figure}[h!]
    % \vspace{-2mm}
    \centering
    \begin{subfigure}[b]{0.32\linewidth}
        \centering
        \includegraphics[width=\linewidth]{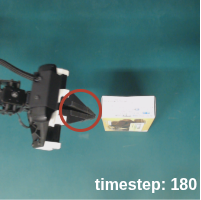}
        \caption{Too Left}
    \end{subfigure}
    \hfill
    \begin{subfigure}[b]{0.32\linewidth}
        \centering
        \includegraphics[width=\linewidth]{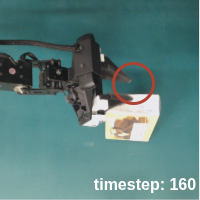}
        \caption{Too Far}
    \end{subfigure}
    \hfill
    \begin{subfigure}[b]{0.32\linewidth}
        \centering
        \includegraphics[width=\linewidth]{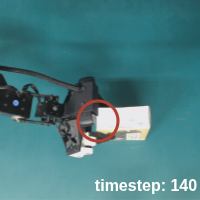}
        \caption{Too Low}
    \end{subfigure}

    \caption{Representative close-up views of manipulation outcomes under execution perturbations. Spatial errors such as \emph{too left} and \emph{too far} lead to task failures, whereas the \emph{too low} case still results in successful grasping execution.}
    \label{fig:failure_examples}
    % \vspace{-4mm}
\end{figure}

Qualitative results show that ORPA preserves ACT's smooth, temporally consistent behavior while enabling corrective adaptation during deployment. ORPA corrected grasping and placement errors in snack box transfer, improved stability in the contact-rich chip tube opening task, and remained robust across object configurations in sorting. Real-world failures also included timing, contact, and object-dynamics errors beyond simple translation and rotation. By predicting residual corrections directly in joint space, ORPA recovered from diverse failures while preserving the base ACT behavior. Corrective feedback was not always required; as shown in Figure~\ref{fig:failure_examples}, the \textit{too low} perturbation remained within ACT's original recovery capability.

\section{CONCLUSION}
In this work, we presented Online Residual Policy Adaptation (ORPA), a lightweight framework for real-time corrective adaptation in robotic manipulation control through human feedback. By augmenting a pretrained control policy with a feedback-conditioned residual module, ORPA enables immediate action-level refinement without requiring retraining of the underlying policy. Unlike rule-based IK correction approaches, the proposed method learns context-aware and temporally consistent joint-space control adjustments that improve robustness under small execution perturbations and distribution shifts. Experimental results on precision-sensitive ALOHA tasks show improved recovery and success rates over baseline ACT and IK-based correction methods, highlighting online residual adaptation as a practical approach for more adaptive robot manipulation.

\section*{Acknowledgment}

We express our sincere gratitude to the National Institute of Advanced Industrial Science and Technology (AIST) for their invaluable support and resources that made this research possible.

\end{document}